\documentclass{article}

\usepackage[preprint]{neurips_2024}

\usepackage[utf8]{inputenc} 
\usepackage[T1]{fontenc}    
\usepackage{hyperref}       
\usepackage{url}            
\usepackage{booktabs}       
\usepackage{amsfonts}       
\usepackage{nicefrac}       
\usepackage{microtype}      
\usepackage{xcolor}         
\usepackage{graphicx}
\usepackage{subcaption}
\usepackage{color}
\usepackage{wrapfig}
\usepackage{amsmath}
\usepackage{multirow}
\usepackage{gensymb}
\newcommand{\yr}[1]{\textcolor{black}{#1}}
\newcommand{\cyg}[1]{\textcolor{black}{#1}}
\newcommand{\hut}[1]{\textcolor{black}{#1}}
\newcommand{\tyz}[1]{\textcolor{black}{#1}}

\title{Plasticine3D: 3D Non-Rigid Editing with Text Guidance by Multi-View Embedding Optimization}

\author{%
  Yige Chen\\
  Shanghai Jiaotong University\\
  \texttt{foxie\_arctic@sjtu.edu.cn} \\
  \And
  Teng Hu\\
  Shanghai Jiaotong University\\
  \And
  Yizhe Tang\\
  Shanghai Jiaotong University\\
  \And
  Siyuan Chen\\
  Shanghai Jiaotong University\\
  \And
  Ang Chen\\
  Shanghai Jiaotong University\\
  \And
  Ran Yi\\
  Shanghai Jiaotong University\\
}

\begin{document}

\maketitle

\begin{abstract}

With the help of Score Distillation Sampling (SDS) and the rapid development of neural 3D representations, some methods have been proposed to perform 3D editing such as adding additional geometries, or overwriting textures. However, generalized 3D non-rigid editing task, which requires changing both the structure (posture or composition) and appearance (texture) of the original object, remains to be challenging in 3D editing field. In this paper, we propose \textbf{Plasticine3D}, a novel text-guided fine-grained controlled 3D editing pipeline that can perform 3D non-rigid editing with large structure deformations. Our work divides the editing process into a geometry editing stage and a texture editing stage to achieve separate control of structure and appearance. In order to maintain the details of the original object from different viewpoints, we propose a {\bf Multi-View-Embedding (MVE) Optimization} strategy to ensure that the guidance model learns the features of the original object from various viewpoints. For the purpose of fine-grained control, we propose {\bf Embedding-Fusion (EF)} to blend the original characteristics with the editing objectives in the embedding space, and control the extent of editing by adjusting the fusion rate. Furthermore, in order to address the issue of gradual loss of details during the generation process under high editing intensity, as well as the problem of insignificant editing effects in some scenarios, we propose {\bf Score Projection Sampling (SPS)} as a replacement of score distillation sampling, which introduces additional optimization phases for editing target enhancement and original detail maintenance, leading to better editing quality. Extensive experiments demonstrate the effectiveness of our method on 3D non-rigid editing tasks.
\end{abstract}


\begin{figure}
    \centering
    \includegraphics[width=0.85\textwidth]{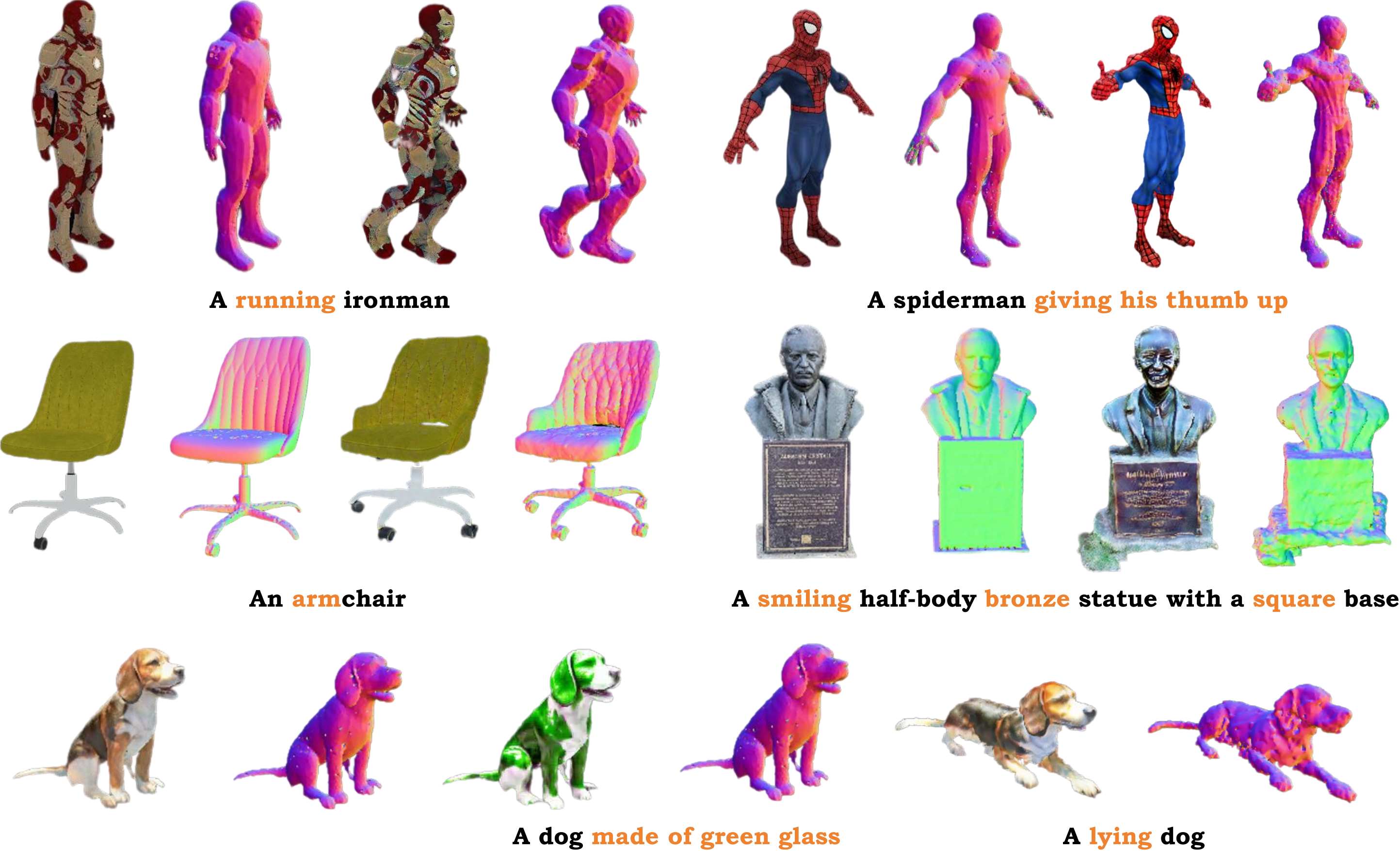}
    \caption{\yr{Our} \textit{Plasticine3D} \yr{achieves text-guided fine-grained controlled 3D editing and enables non-rigid editing with large deformations}. For each case, the original object's RGB and normal maps are displayed on the left side, with the target prompt below, highlighting the main editing objective in orange. The final edited result's RGB and normal maps \yr{are on the right}.}
    \vspace{-0.2in}
    \label{fig:presentation}
\end{figure}

\section{Introduction}


\hut{Recent \yr{developments} in \yr{text-to-3D generation}, particularly those employing score distillation sampling (SDS) and leveraging both 2D priors from text-to-image diffusion models and the inductive bias of implicit 3D representation, have demonstrated promising capabilities in semantic-driven 3D object generation (\cite{poole2022dreamfusion,wang2023score,chen2023fantasia3d}).}
Besides generating 3D objects by text from scratch, there is also \yr{a great demand} in generating new 3D models from existing 3D assets. 
\cyg{Consider\yr{ing} modeling professionals working in the gaming industry have already created a usable 3D model (such as a statue in a game),}
\cyg{they \yr{often need} to create another 3D model with different poses or attire.} 
\hut{\yr{Developing} a 3D generation \yr{method} capable of editing based on existing \yr{3D} models could substantially alleviate the workload for these professionals, allowing them to focus on carving the details.}
\hut{Given the success of SDS methods in generating 3D geometry,}
it immediately raises the question of whether a similar approach could be developed to edit 3D shapes. 

Recent semantic-driven 3D edit\hut{ing} \yr{techniques} can be categorized \yr{into} 3D rigid and 3D non-rigid editing \yr{methods}. 
1) {\bf 3D Rigid Editing methods} perform style transfer tasks on appearance of the object without changing \yr{the structure} (\cite{chiang2022stylizing,fan2022unified,huang2022stylizednerf,zhang2022arf}), 
\yr{which limits their applications.}
2) {\bf 3D Non-rigid Editing methods} \cyg{additionally perform structure \yr{(posture or composition)} changes to the original object besides changing its appearance}. 
\cyg{The existing methods for performing 3D non-rigid editing either rely on a self-generated/provided attention mask/grid to accurately 
\yr{locate a local editing region}
(\cite{zhuang2023dreameditor,sella2023vox,mikaeili2023sked,li2023focaldreamer}), or simply finetune a \hut{Text-to-Image} generative model as the guidance for SDS (\cite{raj2023dreambooth3d}). 
However, these methods either fail to handle significant structural changes, or exhibit poor quality as well as \yr{poor} identity maintenance during the process of structural change.} 
\cyg{We show the comparison between different methods in \yr{Table~}\ref{tab:compare_table}.} 
\hut{In summary, \yr{the} existing 3D editing methods cannot perform high-quality controllable 3D non-rigid editing tasks involving significant structural deformation\yr{s}.} 

To address 
\yr{the above issues,}
we propose \textit{Plasticine3D}, a novel semantic-driven fine-grained controlled 3D editing pipeline that can perform 3D rigid and non-rigid editing \yr{with large structure deformations}. 
For separately controlling the structure and appearance of the 3D object, we divide the editing into 2 stages for geometry and appearance \yr{editing respectively}. 
\cyg{In order to preserve original details of an object from different view\yr{points},} we propose a {\bf Multi-View-Embedding (MVE) Optimization} strategy. 
\cyg{In order to \yr{achieve} fine-grained control of editing extent, we propose {\bf Embedding-Fusion (EF)} to blend the original characteristics with the editing objectives \yr{in the embedding space}.} 
\cyg{In order to strengthen either original \yr{details} or editing target for achieving better editing quality,} we propose {\bf Score Projection Sampling (SPS)} as a replacement for score distillation sampling (SDS) in geometry editing stage. 
Furthermore, in order to further mitigate the impact of the Janus problem, we additionally employ a multi-view normal-depth diffusion model as guidance during the geometry editing stage. 
\hut{Extensive experiments demonstrate the effectiveness of our \textit{Plasticine3D} in both  3D rigid and non-rigid editing tasks.} 

\hut{In summary, our key contributions can be summarized into the following three \yr{aspects}:}
\hut{
\textbf{\textit{1)}} We \yr{propose} \textit{Plasticine3D}, a novel semantic-driven fine-grained controlled 3D pipeline, which can achieve general 3D non-rigid editing \yr{with large structure deformations} while keeping the \yr{the details of the original object}.}
\hut{
\textbf{\textit{2)}} We propose a \textit{Multi-View Embedding (MVE) Optimization} strategy based on perspective-appearance disentanglement, which can extract multi-view features with fine-grained details.}
\hut{
\textbf{\textit{3)}} We design a novel \textit{Score Projection Sampling (SPS)},  
\yr{which introduces additional optimization phases to} focus on structure editing and \yr{original detail} maintenance, largely improving the editing accuracy and \yr{original detail} consistency during 3D object editing.}



\begin{table}
  \caption{Comparison of different 3D editing methods. We categorize 3D editing into: \yr{1) 3D R}igid Editing requiring no structural changes, \yr{2) 3D N}on-Rigid Editing (Composition) requiring simple addition of new objects, and \yr{3) 3D N}on-Rigid Editing (Posture) requiring large deformation on the original structure. $\bigcirc$ means the ability of this method on performing this task is uncertain.}
  \label{tab:compare_table}
  \centering
  \resizebox{0.7\linewidth}{!}{
  \begin{tabular}{lccc}
    \toprule
    \multirow{2}{*}{Method} & \multirow{2}{*}{3D Rigid Editing} & \multicolumn{2}{c}{3D Non-Rigid Editing}  \\
    \cmidrule(r){3-4}
    &   & Composition  & Posture\\
    \midrule
    VOX-E (\cite{sella2023vox}) & \checkmark & $\times$ & $\bigcirc$ \\
    Fantasia3D (\cite{chen2023fantasia3d}) & \checkmark & $\times$ & $\times$ \\
    FocalDreamer (\cite{li2023focaldreamer})  & \checkmark & \checkmark & $\times$ \\
    DreamBooth3D (\cite{raj2023dreambooth3d}) & \checkmark & \checkmark & \checkmark \\
    \textbf{Plasticine3D (Ours)} & \checkmark & \checkmark & \checkmark \\
    \bottomrule
  \end{tabular}}
  \vspace{-0.2in}
\end{table}

\section{Related Works}
\label{sec:related}

\noindent\textbf{Text-Guided Image Editing Methods.}
\hut{Significant advancements in Text-to-Image (T2I) synthesis using diffusion models} have become increasingly conspicuous in recent years. Recently, they have been applied to varies of image editing fields such as inpainting (\cite{lugmayr2022repaint,Avrahami_2022_CVPR}), 2D non-rigid editing (\cite{kawar2023imagic,hertz2022prompt,huang2023kv}) and 2D rigid editing (\cite{yang2023zero, zhang2023adding}). The most relevant fields to our work are non-rigid and rigid editing. 2D rigid editing is a task which only requires changing the appearance of an image while preserving its structure. 
Conversely, 2D non-rigid editing involves modifying both the appearance and structure of an image. 
\hut{Among all, Imagic (\cite{kawar2023imagic}) is one of the most representative 2D non-rigid editing methods which introduces an embedding optimization and fine-tuning process to realize controllable 2D non-rigid editing, inspiring us to explore the realm of 3D non-rigid editing.}

\noindent\textbf{Semantic-Driven 3D Editing Methods.}
Recently, thanks to the rapid development of 2D lifted 3D generative methods that leverage score distillation sampling (SDS) technique, semantic-driven 3D editing has emerged as a promising avenue with significant potential. DreamEditor (\cite{zhuang2023dreameditor}) automatically creates an explicit 3D mask from the attention map with diffusion model to restrict the local edit area. Vox-E (\cite{sella2023vox}) uses cross-attention volumetric grids to refine 
\hut{the editing region}. SKED (\cite{mikaeili2023sked}) performs local editing on 3D scenes by providing precise multi-view sketches. FocalDreamer (\cite{li2023focaldreamer}) possesses the capability of 3D local editing by restricting the target part in a focal region. However, these methods can only perform local editing in a restricted region provided. DreamBooth3D (\cite{raj2023dreambooth3d}) has the ability to perform 3D non-rigid editing while preserving the features from the original object leveraging the \hut{customization} ability of DreamBooth (\cite{Ruiz_2023_CVPR}). However, it suffers from low visual quality results and 
\hut{lacks precise control over editing extents.} 
\hut{In summary, all the existing methods either achieve only local non-rigid editing or exhibit low editing quality.}
In contrast, our work performs high-resolution, fine-grained controlled, semantic-driven 3D non-rigid editing with great flexibility and accurate preservation of the original object \yr{details}.

\begin{figure}[t]
  \centering
  \includegraphics[width=0.9\textwidth]{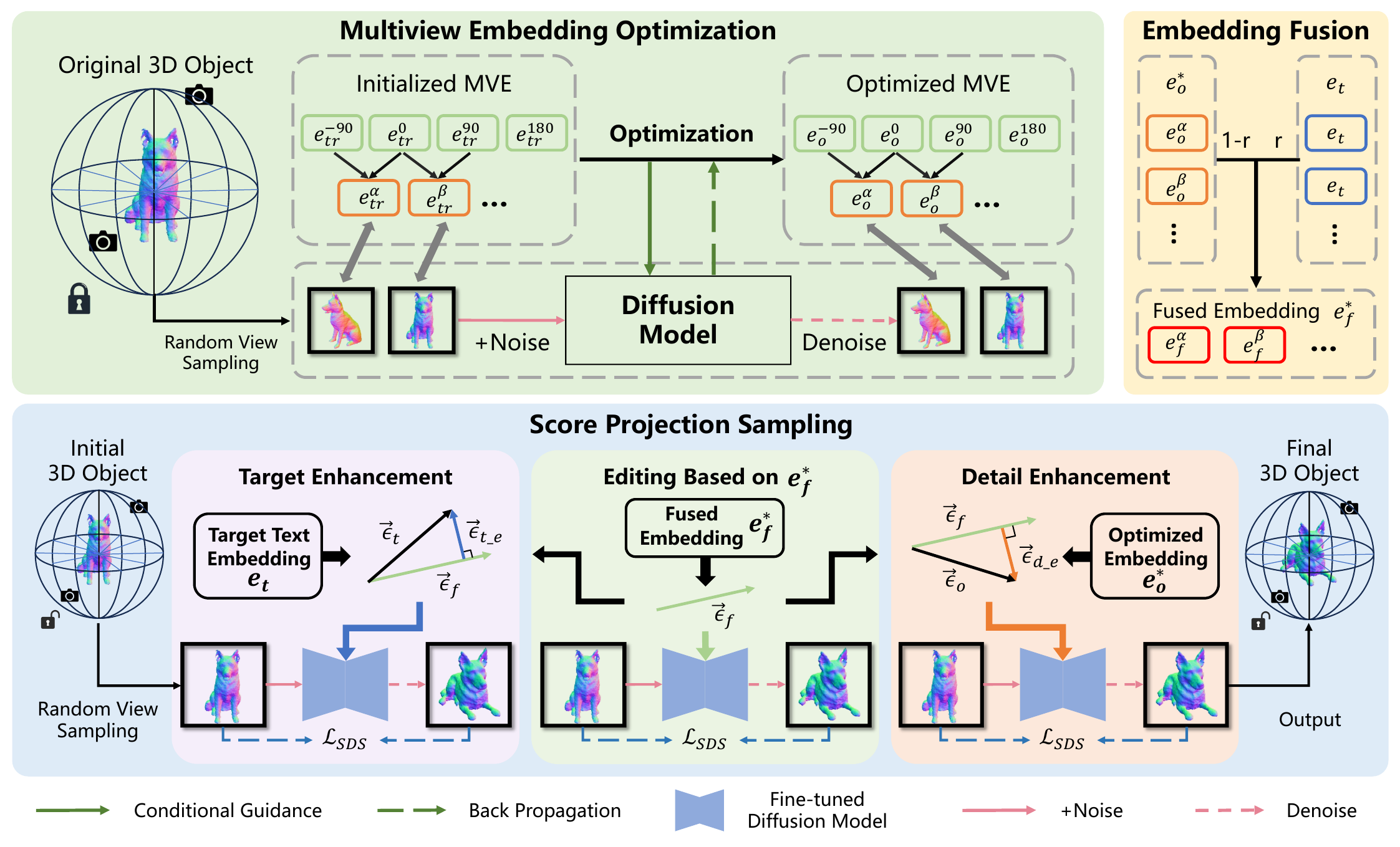}
  \caption{Framework of \textit{Plasticine3D}\cyg{. We first optimize a set of trainable Multi-View Embeddings (MVE) using the rendered images of the original object from \yr{multiple} views \yr{to capture the features of the object}. We also finetune \yr{a Stable} Diffusion model 
  \yr{based on the renderings of the original object.}
  Then we fuse the optimized MVE with the target \hut{text} embedding to get fused embeddings as the semantic guidance in our editing. Finally, we edit the object 
  with our novel three-phase Score Projection Sampling (SPS), \hut{which enhances both the \yr{editing} target features and source details.}}} 
  \label{fig:pipeline}
  \vspace{-0.15in}
\end{figure}

\section{Preliminaries}
\label{sec:preliminary}

\noindent\textbf{Score Distillation Sampling (SDS)}
is a method proposed \yr{in} DreamFusion (\cite{poole2022dreamfusion}) for \yr{text-to-}3D object generation task, \yr{that} distills the hidden priors from \yr{text-to-image} diffusion models to a differentiable 3D representation. 
Given a differentiable 3D representation $x=g(\theta,v)$, where $\theta$ \yr{represents the} trainable parameters, we can obtain the rendered image $x$ from a given view point $v$. 
Subsequently, a sampled noise $\epsilon$ is added to $x$ to a random time step $t$ to get a noisy image $x_t$.
SDS then utilizes a noise predictor $\hat{\epsilon}_{\phi}(x_t;y,t)$ to get a score function from the noisy image $x_t$ given the text embedding $y$ and the noise step $t$. 
\hut{This score function guides the direction of the gradient for updating the parameters $\theta$ of the 3D representation as follows:}
\begin{equation}
\nabla_{\theta}\mathcal{L}_{SDS}(\phi, x) = \mathbb{E}_{t, \epsilon, v}\left[\omega(t)(\hat{\epsilon}_{\phi}(x_t; y, t) - \epsilon)\frac{\partial x}{\partial \theta}\right],
\label{eq:1}
\end{equation}
where $\omega(t)$ is a weight factor related with the time step $t$.
Specifically, $\hat{\epsilon}_{\phi}(x_t; y, t)$ can be calculated using the classifier free guidance used in Fantasia3D \yr{(\cite{chen2023fantasia3d})}:  
\begin{equation}
\hat{\epsilon}_{\phi}(y) = \hat{\epsilon}_{\phi}(e_{unc}) + w * (\hat{\epsilon}_{\phi}(e_{cond}) - \hat{\epsilon}_{\phi}(e_{unc})),
\label{eq:2}
\end{equation}
where $e_{cond}$ refers to a target text embedding generated from the target prompt\yr{,} while $e_{unc}$ refers to an unconditional text embedding generated from an empty string. $x_t$ and $t$ are omitted in the above equation. \hut{For ease of notation, we designate \yr{the direction} $\hat{\epsilon}_{\phi}(e_{cond}) - \hat{\epsilon}_{\phi}(e_{unc})$ as $\vec{\epsilon}_{cond}$. }



\section{Plasticine3D: 3D Non-Rigid Editing Method}
\label{sec:method}

We define our 3D editing task as the following formulation: \yr{G}iven an original 3D object $\Omega_o$ \yr{(represented as} a geometry mesh $M_o$ and a texture $T_o$\yr{)}, along with a target prompt $c_t$ that describes our 
\hut{editing objective}, our goal is to \yr{generate} an edited 3D object $\Omega_e$ \yr{(represented as} a mesh-texture pair $(M_e, T_e)$) \yr{in a way that matches the semantics in the target prompt, while preserving a maximal amount of details of the original 3D object}.
\yr{The 3D editing task can be divided into two categories, 3D non-rigid editing and rigid editing, based on whether the structure is changed:} 
\yr{1)} 3D rigid editing \yr{only} requires changing the appearance (texture) of an object, while preserving its structure (posture or composition);
while \yr{2)} 3D non-rigid editing requires changing both the structure (posture or composition) of an object and the appearance (texture).

\cyg{Compared to rigid editing of 3D content, \yr{3D} non-rigid editing requires greater editing flexibility and larger overall deformations. 
}
\yr{Therefore, although the existing methods can successfully perform 3D rigid editing (\cite{chiang2022stylizing,fan2022unified,huang2022stylizednerf,zhang2022arf}),} \cyg{or 3D non-rigid editing that simply requires adding an object to the original structure (\cite{zhuang2023dreameditor,li2023focaldreamer})}, 
\yr{they often cannot handle 3D non-rigid editing with large deformations. 
When performing non-rigid editing that requires large structure changes, these existing methods often lose part of the features of the original model.} 
\yr{To address this issue, we aim to develop a method to} perform high-quality 3D non-rigid editing, \yr{which enables flexible editing with large deformations}.

\hut{We employ \yr{a} 2-stage editing pipeline} for non-rigid editing task, \yr{where} geometry and texture \yr{editing is performed separately in each stage}. 
This 2-stage pipeline also gives us enough flexibility when the texture $T_o$ is not provided or when we need to separately adjust the edited geometry and texture.

\cyg{
\yr{In} each stage, \yr{we design three steps to achieve the desired editing (Fig.~\ref{fig:pipeline}): Multi-View-Embedding Optimization, Embedding Fusion, and Score Projection Sampling.}
\yr{1)} We first optimize a Multi-View Embedding (MVE) based on viewpoint-appearance decoupling\yr{, optimizing different embeddings for different views, which} ensures that the guidance model learns the features of the object from various viewpoints. 
\yr{2)} Then, we fuse the optimized embeddings with the target embeddings to integrate information through Embedding Fusion (EF), \yr{which} ensures the fine-grained control of editing extent through embedding fusion rate. 
\yr{3)} Finally, we \yr{propose} Score Projection Sampling (SPS) to update an initial mesh for high-quality 3D non-rigid editing, \yr{which} can enhance the quality of editing by using vector augmentation during the editing process.
}

\subsection{Multi-View-Embedding (MVE) Optimization}
\label{subsec:MVE}

\cyg{
\yr{An important aspect}
in 3D non-rigid editing task} is to maintain the \yr{detail} information of the original \yr{3D} object \yr{while performing large deformations}. 
\cyg{\yr{In order to capture the features of the original object,} inspired by Imagic (\cite{kawar2023imagic}), a 2D image editing method, we propose a strategy \yr{that optimizes} both the text embedding and the diffusion model 
\yr{to reconstruct the multi-view information of the original 3D object.}} 
\yr{Our key observation is that:} Different from \yr{the} 2D scenario\yr{s}, a single trainable text embedding in \yr{each editing} stage is not enough for keeping \yr{the detail} information of the original object. 
\yr{This is because,}
unlike 2D images, each rendering angle of 3D objects corresponds to a different image, 
\yr{which means that} renderings of a 3D object from different perspectives correspond to different embeddings in CLIP Space. 
\yr{Therefore}, unlike 2D images, different trainable embeddings are needed for different perspectives of the same \yr{3D} object. 
\yr{Motivated by this observation}, we propose Multi-View Embedding (MVE) Optimization, a \yr{novel} strategy for preserving the \yr{details} of the original \yr{3D} object 
\yr{by optimizing different embeddings for different views.}

\noindent\textbf{Multi-View Embedding.}
\cyg{In this step, we \yr{aim to obtain an} optimized text embedding $e_o$ that best matches the original \yr{3D} object semantically. 
We first obtain a target text embedding $e_t$ \yr{from} the \yr{target} text prompt $c_t$ \yr{using} the CLIP encoder\yr{, and initialize the trainable embedding $e_{tr}$ of the original 3D object as a clone of $e_t$.}
Then we \yr{render the original 3D object from multiple viewpoints, and feed the renderings into a pretrained Stable Diffusion model.} \yr{The trainable} embedding $e_{tr}$ \yr{is then optimized} using the following reconstruction loss:}
\begin{equation}
\mathcal{L}_{recon} = \mathbb{E}_{v, t, \epsilon}\left[\Vert{\epsilon}_{\phi}(x^v_t; e_{tr}, t) - \epsilon\Vert^2_2\right],
\label{eq:3}
\end{equation}
where $v$ is a randomly selected view\yr{point}, $x^v$ is a rendered image from the selected viewpoint of the original 3D object $\Omega_o$, and $x^v_t$ is the noised version of $x^v$. ${\epsilon}_{\phi}(x^v_t; e_{tr}, t)$ is the noise prediction given by the UNet inside diffusion model, $\epsilon\sim N(0,1)$, $t\sim Uniform(1,T)$. 


\cyg{However, as we stated above, \yr{since a 3D object has different appearances under different views, a single text embedding is not sufficient to represent the 3D object, \textit{i.e.,}} different trainable embeddings are needed for different perspectives of the same \yr{3D} object.} \hut{A direct solution is to assign $n_v$ independent trainable embeddings for $n_v$ different perspectives, with each embedding covering $\frac{360}{n_v}$ degrees. But \yr{this solution will cost} a significant amount of computation resources as $n_v$ increases.}
\yr{In this paper,} we \yr{propose} an \textit{interpolation-based} \cyg{multi-view} \yr{trainable embedding} as our final solution. 
We \yr{first set} four azimuth angles as the \textit{base azimuth angles}: 0\degree, -90\degree, 90\degree, and 180\degree (equivalent to -180\degree).
We \yr{then} assign a trainable text embedding to each of the four azimuth angles. 
For any azimuth angle $\alpha$ in the range \yr{of} $[-180\degree,180\degree]$, its corresponding embedding is the interpolation of the trainable embeddings at its adjacent two base azimuth angles:
\begin{equation}
e^{\alpha}_{tr} = \frac{\alpha - \alpha_0}{90} * e^{\alpha_0}_{tr} + (1 - \frac{\alpha - \alpha_0}{90}) * e^{\alpha_1}_{tr},
\label{eq:4}
\end{equation}
where $e^{\alpha}_{tr}$ \yr{is} the corresponding embedding at azimuth angle $\alpha$, and $\alpha_0$, $\alpha_1$ are two base azimuth angles adjacent to $\alpha$. 
\tyz{We \yr{use the interpolated view-dependent embedding $e^{\alpha}_{tr}$ in} reconstruction loss:}
\begin{equation}
\mathcal{L}_{recon} = \mathbb{E}_{v, t, \epsilon}\left[\Vert{\epsilon}_{\phi}(x^v_t; e^{\alpha}_{tr}, t) - \epsilon\Vert^2_2\right],
\label{eq:5}
\end{equation}
where 
\yr{$\alpha$ is the corresponding azimuth angle of viewpoint $v$.}
Since the \cyg{multi-view} embeddings \yr{are calculated by} linear interpolation, \yr{they} \cyg{are} fully differentiable with respect to the base embedding\yr{s. This} ensures that \yr{they can be optimized} through gradient propagation. 
\cyg{After the trainable multi-view embeddings $e^*_{tr}$ \yr{have} been fully optimized, we take their final state as the optimized multi-view embeddings $e^*_o$\yr{, which} correspond to the multi-view features of the original object in the text space.}

\noindent\textbf{UNet Finetuning.}
\cyg{While the optimized multi-view embeddings $e^*_o$ are the embeddings that best matches \yr{the original 3D object} $\Omega_o$, there is still a difference in details between $\Omega_o$ and 
\hut{the \yr{3D} representation captured by $e^*_o$}. 
\hut{This discrepancy arises due to \yr{that} text-space embeddings serv\yr{e} as relatively global feature representations, thereby struggling to encapsulate the intricate details of the original \yr{object}.} 
This limitation has been confirmed in some diffusion-based image customization methods (\cite{ye2023ip-adapter,li2023photomaker}). 
Therefore, 
\hut{it arises the necessity to further finetune the} UNet architecture \yr{in the pretrained Stable Diffusion} and \hut{enhance the alignment between} the optimized multi-view embeddings $e^*_o$ and the detailed information of the original \yr{3D} object $\Omega_o$.} 
We employ the same reconstruction loss \hut{\yr{as} in Eq.(\ref{eq:5}) for} this finetuning process, but \yr{optimizes the UNet parameters instead of the trainable embeddings,} 
\yr{\textit{i.e.,} we keep the embeddings fixed as the optimized multi-view embeddings $e^*_o$.}

\subsection{Embedding Fusion (EF)}

From MVE optimization, we \yr{obtain} a set of multi-view optimized embeddings $e^*_o$, along with a fine-tuned diffusion model whose parameters are represented as $\phi$. 
In order to achieve 
\yr{fine-grained editing with explicit control of editing extents,} 
we blend the optimized multi-view embeddings $e^*_o$ (corresponding to the original identity features) with the target embeddings $e_t$ (corresponding to the editing objective) by \yr{linear interpolation:} 
\begin{equation}
e^*_f = r * e_t + (1-r) * e^*_o,
\label{eq:6}
\end{equation}
where $r$ \yr{represents} the embedding fusion rate \yr{within $[0,1]$. In this way, we edit towards the target prompt while preserve the features of the original object, and the balancing between these two aspects can be adjusted by the fusion rate.} After we get the fused embeddings $e^*_f$, it will be used in \yr{our} Score 
\yr{Projection Sampling} as a conditional guidance to generate an edited 3D object. \yr{Experiments} in \yr{Sec.~}\ref{EF_results} \yr{show} that the granularity of the editing results \yr{can be adjusted} by the rate $r$.

\subsection{Score Projection Sampling (SPS)}
\label{sec:SPS}

\yr{After obtaining the fused embedding $e^*_f$, which is a blended version of editing target and optimized multi-view embeddings, we aim to conduct 3D editing guided by $e^*_f$\footnote{\yr{Compared to editing by target embedding $e_t$ directly, editing by $e^*_f$ enables fine-grained control of editing extents, \textit{i.e.,} users can control the granularity of editing by fusion rate $r$ ($r=1$ is the case of editing by $e_t$).}}.
A na\"ive solution is to employ the Score Distillation Sampling (SDS) loss (Sec.~\ref{sec:preliminary}) widely used in text-to-3D generation to optimize a 3D mesh, using $e^*_f$ as the conditional guidance, with the original 3D object or a base ellipsoid as the initial shape.}
\cyg{However, as observed in both 2D and 3D scenarios, embedding fusion-based methods have inherent drawbacks. 
One major issue is that as the embedding fusion rate $r$ increases, since the embeddings \yr{of the original object} are blended less to the fused embeddings, the \yr{features of the} original \yr{object} gradually diminish. 
\yr{Another issue is that when using the SDS loss to update from the original 3D object based on $e^*_f$, it is often difficult to achieve large deformation, since the diffusion model has been finetuned on the original 3D object.}
}


\cyg{To address these challenges, inspired by the Perp-Neg algorithm (\cite{armandpour2023re}), we \yr{propose} \textit{Score Projection Sampling (SPS)}, a modified version of SDS, to mitigate these issues.}
\yr{We mainly add two additional optimization phases on the basis of SDS editing based on fused embedding, with a target enhancement phase at the beginning, and a detail enhancement phase at the end.}
SPS is divided into \yr{three phases:} 
\yr{Phase 1: \textit{Target Enhancement} phase} first strengthens the guidance of the editing target to generate features with the desired modifications in the object. 
\cyg{Phase 2: \textit{Editing based on fused embedding} aligns the optimization direction with the editing target described by the fused embedding. \yr{Note that the na\"ive solution only has this phase.}
\yr{Phase 3: \textit{Detail Enhancement} phase:} Once the editing target is sufficiently achieved, it then reinforces the guidance of the original details to \yr{maintain} the original object's features.
}


\noindent\textbf{Target Enhancement.}
\cyg{In this \yr{phase}, our goal is to enhance the guidance of the editing target based on SDS to generate features with the desired 
transformations \hut{while not affecting the guidance direction $\vec{\epsilon}_f$ derived from the fused embeddings}. 
}

\cyg{\yr{Given a target vector $\vec{\epsilon}_t$ derived from the target embedding $e_t$ by $\hat{\epsilon}_{\phi}(e_t) - \hat{\epsilon}_{\phi}(e_{unc})$,} the most \tyz{direct} \yr{way} is to add $\vec{\epsilon}_t$ to the guiding vector $\vec{\epsilon}_f$. However, as mentioned in Perp-Neg, directly adding this vector may affect the strength of the fused guiding direction \yr{$\vec{\epsilon}_f$}. Therefore, as shown in Fig.~\ref{fig:pipeline}, we extract a 
component of the target vector \yr{$\vec{\epsilon}_t$} perpendicular to the fused vector \yr{$\vec{\epsilon}_f$, denoted as $\vec{\epsilon}_{t\_e}$}, and add \yr{a weighted version of} it to the fused vector \hut{to obtain the final target-enhancement direction $\vec{\epsilon}$}:}
\begin{equation}
    \begin{aligned}
        \vec{\epsilon}_{t\_e} = \vec{\epsilon}_{t} - \frac{\vec{\epsilon}_{t} \cdot \vec{\epsilon}_{f}}{\Vert\vec{\epsilon}_{f}\Vert^2}\vec{\epsilon}_{f}, \quad \vec{\epsilon}=\lambda_{t}\vec{\epsilon}_{t\_e}+\vec{\epsilon}_{f}.
    \end{aligned}
    \label{eq:7}
\end{equation}
\hut{The perpendicular component \yr{$\vec{\epsilon}_{t\_e}$} reinforces guidance toward the target, while minimally altering the guidance direction obtained from the fused embeddings.
We name it the \textit{target-enhancement direction}.} By weighting and incorporating this direction into the CFG noise prediction,
SDS optimization can better reflect the editing objectives \yr{and help to achieve large deformations}. 


\noindent\textbf{Editing Based on Fused Embedding.}
\cyg{SPS is essentially an \yr{improvement based on} SDS guided by a fused embedding $e^*_f$ as the editing condition. 
During the initial and final steps of SDS optimization process, we add Target Enhancement and Detail Enhancement \yr{phases} respectively, to achieve the large deformations in 3D non-rigid editing tasks and to maintain the details of the original object. 
However, \yr{apart from target and detail enhancement}, the semantic goal of the editing corresponding to $e^*_f$ still need\yr{s} to be maintained. 
So between these \yr{two} phases, we keep the direction of optimization aligning with the fused direction $\vec{\epsilon}_f$ to ensure that the editing result meet\yr{s} the semantic goal. Therefore, after achieving significant deformations during the target enhancement \yr{phase}, we use $\vec{\epsilon}_f$ to perform the original SDS to maintain the semantic features of the editing target. Once the semantic goals are sufficiently maintained, we proceed to the next \yr{phase} for performing detail optimization.
}

\noindent\textbf{Detail Enhancement.}
\cyg{In this \yr{phase}, our goal is to strengthen the guidance of the original details after the editing target has been sufficiently achieved, in order to restore the original details of the object. During this process, we also aim to ensure that this enhancement does not affect the guidance direction \yr{$\vec{\epsilon}_f$} derived from the fused embeddings.}

\yr{We first obtain a vector $\vec{\epsilon}_o$ for the optimized embedding $e^*_o$ by}
$\hat{\epsilon}_{\phi}(e^*_o) - \hat{\epsilon}_{\phi}(e_{unc})$.
\cyg{Since $e^*_o$ corresponds to the original object, we should steer the guiding direction \yr{$\vec{\epsilon}_{f}$} towards $\vec{\epsilon}_o$ to enhance the preservation of original details.} Similar to the target enhancement term, we perform this enhancement by \yr{extracting} a perpendicular vector \yr{from} $\vec{\epsilon}_f$ to $\vec{\epsilon}_o$\yr{, denoted as $\vec{\epsilon}_{d\_e}$, and adding a weighted version of it to $\vec{\epsilon}_{f}$} \hut{to obtain the final detail-enhancement direction $\vec{\epsilon}$}:
\begin{equation}
    \begin{aligned}
\vec{\epsilon}_{d\_e} = \vec{\epsilon}_{o} - \frac{\vec{\epsilon}_{o} \cdot \vec{\epsilon}_{f}}{\Vert\vec{\epsilon}_{f}\Vert^2}\vec{\epsilon}_{f}, \quad \vec{\epsilon}=\lambda_{d}\vec{\epsilon}_{d\_e}+\vec{\epsilon}_{f}.
\end{aligned}
    \label{eq:8}
\end{equation}
\yr{The} perpendicular component \yr{$\vec{\epsilon}_{d\_e}$} reinforc\yr{es} the guidance towards the details of the original object, while minimally affecting the guidance direction derived from the fused embeddings. We name it the \textit{detail-enhancement \hut{direction}}. By weighting and \hut{incorporating} this term into the CFG noise prediction,  
the SDS optimization can recover more details of the original object after the editing target is achieved.

\subsection{Multi-View Normal-Depth Guidance}
Although \yr{our} multi-view embeddings and \yr{UNet} finetuning enable the model to \yr{learn} features and details of the original object from different perspectives, 
multi-view embeddings \yr{alone cannot} ensure geometric consistency \yr{across different views} in the subsequent generation process.
Therefore, we \yr{further} introduce a multi-view normal-depth diffusion model from RichDreamer~\cite{qiu2023richdreamer}, \cyg{\yr{which is} trained on large-scale 3D dataset}, as a regularization guidance to ensure geometric consistency in the generation process and mitigate the impact of the Janus problem. 

\cyg{\yr{We utilize the multi-view normal-depth diffusion model in the geometry editing stage 
with three steps:}
\yr{Firstly,}
similar to MVE, we also \yr{optimize} a trainable embedding \yr{to capture} the \yr{features} of the original object in the geometric space of the normal-depth diffusion model. 
\yr{Secondly}, we fuse the optimized embedding with the target embedding obtained from the target prompt with the same embedding fusion rate \yr{as} in EF, resulting in a fused embedding on the normal-depth diffusion model. 
\yr{Thirdly,} we incorporate this fused embedding into the overall loss calculation by computing a SDS loss with the normal-depth diffusion model. The weight of this loss is $\frac{1}{8}$ of the weight of SPS loss.}

\begin{figure}[t]
  \centering
  \includegraphics[width=1.0\textwidth]{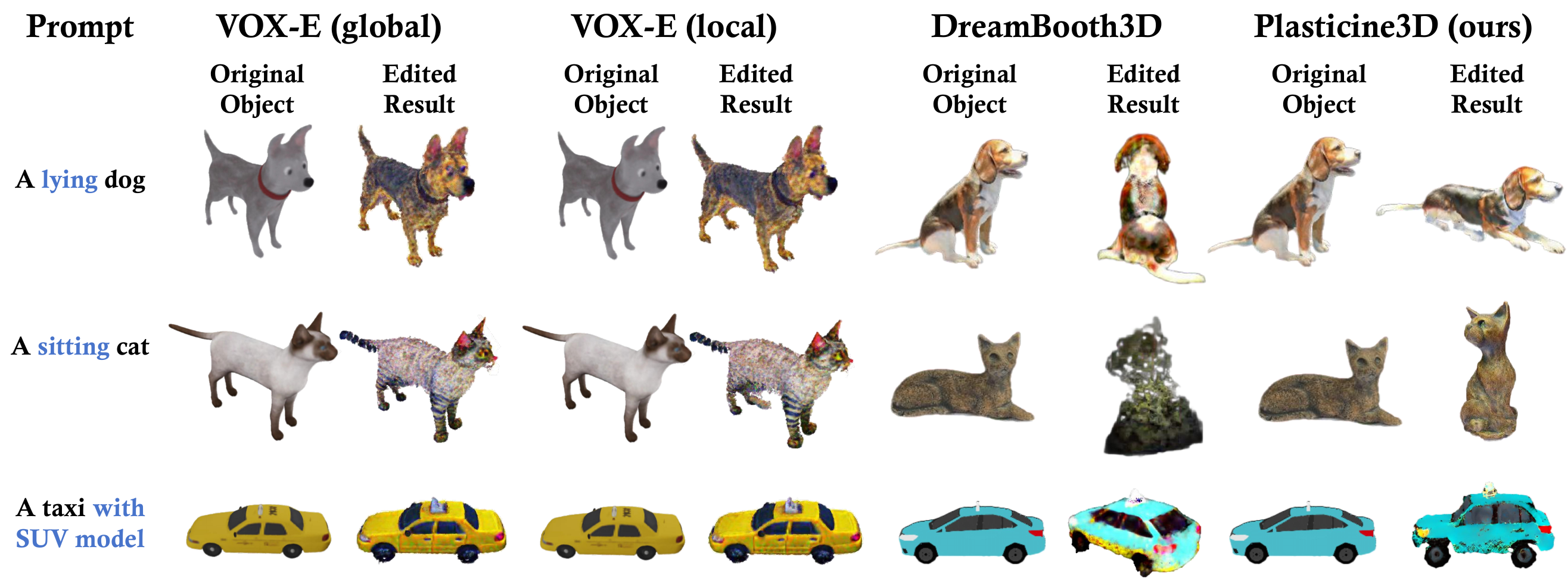}
  \caption{Qualitative comparisons: Vox-E (global) and Vox-E (local) lack the ability in handling 3D non-rigid editing involving large structure deformations. DreamBooth3D has some ability in performing large structure deformations, \yr{but} lacks stability and suffers from low quality. \yr{In contrast, our \textit{Plasticine3D} achieves good performance in editing accuracy and large deformations.}}
  \label{fig:qualitative_comparison}
\end{figure}

\section{Experiments}
\label{sec:experiments}

\subsection{Implementation Details}
\label{subsec:implementation_details}
We use threestudio (\cite{threestudio2023}) as the
\yr{codebase for implementing}
our pipeline. \hut{And we employ} Stable Diffusion \yr{2.1}
as our guidance model for fine-tuning and \yr{SPS}. 
We also use multiview normal-depth diffusion model from RichDreamer (\cite{qiu2023richdreamer}) as a multiview guidance for enhancing geometry consistency. For 3D representation, we use NeuS in geometry editing stage, and DMTeT in texture editing stage with nvdiffrast (\cite{Laine2020diffrast}) as the renderer. 
 \hut{Our original 3D assets are sourced from Objaverse (\cite{deitke2023objaverse}).}
\hut{Experiments are conducted on a single V100-32GB GPU with a batch size of $12$} 
by gradient accumulation. 
More details 
are included in Appendix.

\subsection{Comparison Experiments}
\label{subsec:comparison_experiments}

\noindent\textbf{Baselines.}
We compare \textit{Plasticine3D} with 
\yr{three baselines: Vox-E (global) (\cite{sella2023vox}), Vox-E (local), and DreamBooth3D (\cite{raj2023dreambooth3d}). More details are in Appendix~\ref{sec:baselines}.}

\noindent\textbf{Qualitative \yr{Comparison}.}
The qualitative comparison with 3D editing baselines is shown in \yr{Fig.~\ref{fig:qualitative_comparison}:} Vox-E (global) 
\hut{struggles to handle 3D non-rigid editing tasks involving large structural changes.}
Vox-E (local) can perform 3D non-rigid editing locally, 
\hut{but faces challenges in precisely defining target regions} when the target prompt describes an overall status (shown in \yr{Fig.~\ref{fig:qualitative_comparison}}). 
DreamBooth3D can perform 3D non-rigid editing with relatively large structure changes\yr{, but} 
suffers from instability, geometry inconsistency, and poor editing quality. In contrast, 
\hut{our method consistently achieves high-quality 3D non-rigid editing results that align closely with the editing objectives.}

\begin{wraptable}{r}{0.45\textwidth}
  \caption{Quantitative \yr{comparison with the baselines. Our \textit{Plasticine3D} achieves the best scores on two evaluation metrics.}}
  \label{tab:quantitative_comparison}
  \centering
  \resizebox{\linewidth}{!}{
  \begin{tabular}{lcc}
    \toprule
    Method & $CLIP_{sim}$ $\uparrow$ & $CLIP_{dir}$ $\uparrow$  \\
    \midrule
    VOX-E (global) & 0.285 & 0.0278\\
    VOX-E (local) & 0.286 & 0.0232\\
    DreamBooth3D & 0.250 & 0.0138\\
    Plasticine3D (Ours) & \textbf{0.295} & \textbf{0.0456} \\
    \bottomrule
  \end{tabular}}
\end{wraptable}

\noindent\textbf{Quantitative \yr{Comparison}.}
\yr{The} quantitative comparison result\yr{s are reported} in Tab.~\ref{tab:quantitative_comparison}. We use \yr{two evaluation metrics (details in Appendix~\ref{sec:evaluation_metrics}): 1)} the CLIP similarity $CLIP_{sim}$ \yr{measures} the alignment between the edited result and the target prompt;
\yr{and 2) the} CLIP directional similarity $CLIP_{dir}$ \yr{(\cite{stylegan-nada}) measures} the 
\yr{alignment of CLIP-space direction between the text and rendering image pairs of the source and the edited 3D object.}
\hut{To ensure fairness across all view\yr{points} of a 3D object,}
we calculate the average $CLIP_{sim}$ and $CLIP_{dir}$ \hut{over} 100 rendered images of an object from random \yr{viewpoints} with the target editing prompt to get the final result. 
\yr{We achieve the best score \hut{across} all the \yr{comparison methods on} 3D non-rigid editing tasks.}

\begin{figure}[t]
  \centering
  \includegraphics[width=0.9\textwidth]{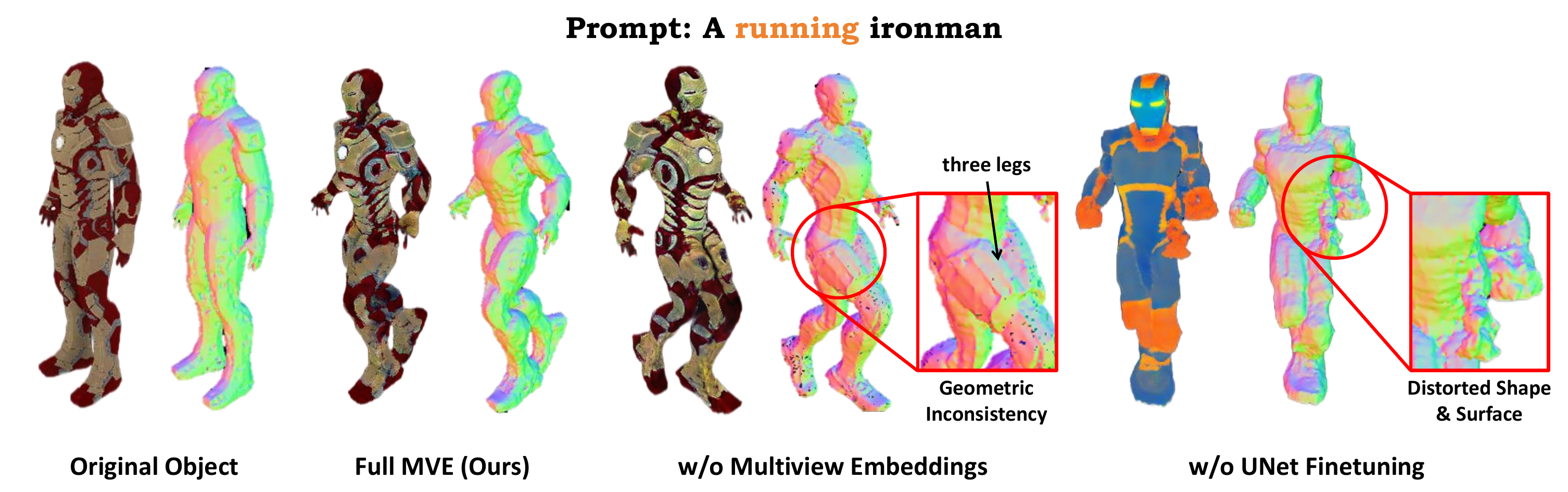}
  \caption{Ablation experiment for components in MVE optimization. Without multi-view embeddings, the editing results suffer from serious geometric inconsistency (three legs). Without UNet finetuning, the editing results are distorted and lose original details.}
  \label{fig:MVE}
\end{figure}

\begin{figure}[t]
  \centering
  \includegraphics[width=0.75\textwidth]{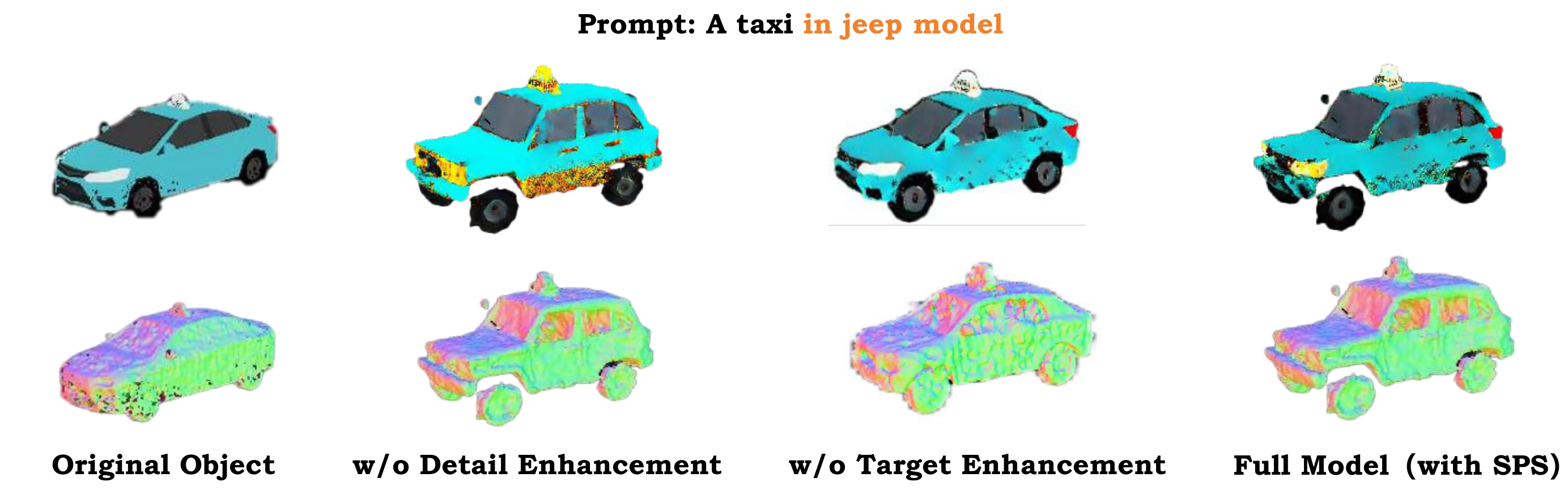}
  \caption{
  \yr{Ablation studies on}
  Score Projection Sampling (SPS). Without target enhancement phase in SPS, the editing results show poor alignment with the editing target. Without detail enhancement phase is SPS, the editing results lose the details of the original object.}
  \label{fig:SPS}
\end{figure}

\subsection{Ablation Studies}

\noindent\textbf{MVE Optimization.} 
As stated in \yr{Sec.~}\ref{subsec:MVE}, \yr{the} Multi-View Embedding Optimization strategy is crucial for ensuring the quality of the editing results. 
In \yr{Fig.~}\ref{fig:MVE}, we 
\yr{conduct ablation studies on MVE optimization}
by disabling components in MVE respectively. Specifically, we compare the editing results of: \yr{1)}  \yr{our full model (with} MVE), \yr{2)} the \yr{ablated model without} the multi-view trainable embedding component and replacing it with a single trainable embedding for editing, and \yr{3)} \yr{the ablated model without} the UNet finetuning step. 
The results show that: without the multi-view trainable embeddings, the editing results not only fail to maintain correspondence between details, 
but also exhibit severe geometric inconsistencies \yr{(\textit{e.g.,} three legs in Fig.~\ref{fig:MVE});} without \yr{the} UNet finetuning step, the details of the \yr{edited} object deviate significantly from those of the original object. 

\noindent\textbf{Score Projection Sampling (SPS).}
As stated in Sec.~\ref{sec:SPS}, our Score Projection Sampling (SPS) strategy is crucial to the quality of editing results. 
In Fig.~\ref{fig:SPS}, we \yr{conduct ablation studies on SPS} by disabling enhancement phases in SPS respectively. Specifically, we compare the editing results of: 1) our full model (with SPS), 2) the \yr{ablated model without} the target enhancement phase, 
and 3) the \yr{ablated model without} the detail enhancement phase. 
The results show that: without \yr{the} target enhancement phase, the editing results fail to perform large-scale deformation \yr{on} the original object\yr{, with the edited structure similar to the original object}; without \yr{the} detail enhancement phase, \yr{the results} fail to align with the details of the original object. 
The \yr{ablation study} shows SPS is important for aligning the editing result with our editing target as well as the details of the original object.

\section{Conclusions}
\label{sec:conclusion}

We propose \textit{Plasticine3D}, a novel \yr{text-guided, fine-grained controlled} 3D editing pipeline \yr{that can perform 3D non-rigid editing with large structure deformations}. Our \yr{method} divides the editing process into a geometry editing stage and a texture editing stage to achieve \yr{separate control of structure and appearance}. We propose a Multi-View-Embedding (MVE) Optimization strategy to preserve the \yr{details} of the original object \yr{from different viewpoints}, and Embedding-Fusion (EF) to \yr{achieve fine-grained} control \yr{of} the extent of editing by adjusting the embedding fusi\yr{on} rate. Furthermore, in order to address the issue of gradual loss of details during the generation process under high editing intensity, as well as the problem of insignificant editing effects in some scenarios, we \yr{propose} Score Projection Sampling (SPS) \yr{as} a replacement of score distillation sampling, which \yr{introduces additional optimization phases for editing target enhancement and original detail maintenance,} lead\yr{ing} to better editing quality. 
\yr{Extensive experiments demonstrate that} our pipeline can perform 3D non-rigid editing \yr{with large structure deformations,} with fine-grained control and \yr{good editing} quality. 

\clearpage


\medskip

\bibliographystyle{abbrvnat}
\bibliography{neurips_2024}





\clearpage
\appendix

\section{Appendix}
\subsection{Overview}
\hut{In this \yr{appendix}, more details about the proposed \textit{Plasticine3D} method and more experimental results are provided, including:}

\begin{itemize}
    \item More implementation details (Section~\ref{sec:more implementation details});
    \item More details for the training details of the proposed MVE (Section~\ref{sec:Training Details of MVE});
    \item \yr{Details of evaluation metrics (Section~\ref{sec:evaluation_metrics});}
    \item \yr{Details of comparison baselines (Section~\ref{sec:baselines});}
    \item Controllable results with Embedding-Fusion (Section~\ref{EF_results});
    \item 
    \yr{Details and} hyperparameters of SPS (Section~\ref{sec:Hyperparameters of SPS});
    \item More experiment results (Section~\ref{sec:more experiment resutls});
    \item Limitations of our work (Section~\ref{sec:limitations});
    \item Potential societal impacts of our work (Section~\ref{sec:Societal Impact}).
    
\end{itemize}

\subsection{More Implementation Details}
\label{sec:more implementation details}
As mentioned in the main \yr{paper}, our method is a 2-stage pipeline. In each stage, our method independently edits the geometry or texture of the original object with MVE, EF and SPS. 1) During the geometry editing stage, we use NeuS as our 3D representation, which enhances flexibility in this stage. Multiview Normal-Depth Guidance \yr{are} also \yr{incorporated} in this stage for enhancing geometric consistency. Additionally, we follow the approach of the 3D generation method Fantasia3D by dividing the geometry optimization into coarse and fine optimization. We optimize NeuS for $3,000$ steps in total. In the coarse optimization stage (first $2,000$ steps), the rendered normal map and opacity are resized and directly input into the finetuned UNet to calculate the SPS loss. In the fine optimization stage (last $1,000$ steps), the normal map is encoded by the VQ-VAE encoder before being sent to the UNet to calculate the SPS loss. Fantasia3D has demonstrated that this method significantly improves the quality of geometry generation. 2) During the texture editing stage, we use DMTeT as our 3D representation.

\subsection{\yr{More} Training Details of MVE}
\label{sec:Training Details of MVE}
We use Stable Diffusion 2.1 as our T2I diffusion model. For Multi-View Embeddings, we employ the Adam optimizer with a learning rate of $2e-3$ and a batch size of $4$ for $1,000$ steps of optimization. For UNet finetuning, we use the AdamW optimizer with a learning rate of $5e-7$ and a batch size of $4$ for $1,500$ steps of optimization. Due to limitations in our GPU memory capacity, we utilize gradient accumulation during the UNet finetuning process to save memory. 

\subsection{Details of Evaluation Metrics}
\label{sec:evaluation_metrics}

\yr{We adopt two evaluation metrics in quantitative comparisons (Sec.~\ref{subsec:comparison_experiments}). In this section, we provide details on the calculation formula for these evaluation metrics.
}

\yr{1)} The CLIP similarity $CLIP_{sim}$ \yr{measures} the alignment between the edited result and the target prompt, calculated \yr{by}:
\[
CLIP_{sim}(I, c)=\max(\cos(E_I, E_c), 0),
\]
which corresponds to the cosine similarity between visual CLIP embedding $E_I$ for an image $I$ and textual CLIP embedding $E_c$ for an caption $c$. In our case, $I$ is a rendered image from the edited object and $c$ is our target prompt $c_t$.

\yr{2) The} CLIP directional similarity $CLIP_{dir}$ \yr{measures} the 
\yr{alignment of CLIP-space direction between the text and rendering image pairs of the source and the edited 3D object,}
calculated by:
\[
CLIP_{dir}(I\_t, I\_o, c\_t, c\_o)=\max(\cos(E_{I\_t} - E_{I\_o}, E_{c\_t} - E_{c\_o}), 0),
\]
\yr{where} $I\_o$ and $I\_t$ are the rendered images from the original object and the edited \yr{object}.
$c\_t$ and $c\_o$ \yr{represent} the editing target prompt, and the caption of the original object.
$E_{I\_o}$ and $E_{I\_t}$ \yr{represent} the visual CLIP embedding\yr{s} of the original image $I\_o$ and the edited image $I\_t$\yr{, respectively}. 
$E_{c\_t}$ and $E_{c\_o}$ \yr{represent the} textual CLIP embedding\yr{s} of the target prompt $c\_t$ the original image caption ${c\_o}$\yr{, respectively}.

\subsection{\yr{Details of Comparison Baselines}}
\label{sec:baselines}

\yr{In the experiments (Sec.~\ref{subsec:comparison_experiments}), we compare with three baseline methods. In this section, we provide more descriptions of these baselines}: 1) Vox-E (global) (\cite{sella2023vox}): a state-of-the-art (SOTA) method for 3D object editing leveraging 2D diffusion model\yr{, which} enables global-scale 3D editing that aligns with a provided target prompt. 
2) Vox-E (local): Vox-E also provide a local mode in its pipeline. Vox-E (local) leverages the attention grid to enable 3D non-rigid editing in local regions. 3) DreamBooth3D (\cite{raj2023dreambooth3d}): a 3D non-rigid editing method based on DreamBooth (\cite{ruiz2023dreambooth}). 
\yr{For Vox-E (global) and Vox-E (local), we use their official codes.}

\subsection{Controllable Results with Embedding-Fusion (EF)}
\label{EF_results}

\begin{figure}[t]
  \centering
  \includegraphics[width=0.8\textwidth]{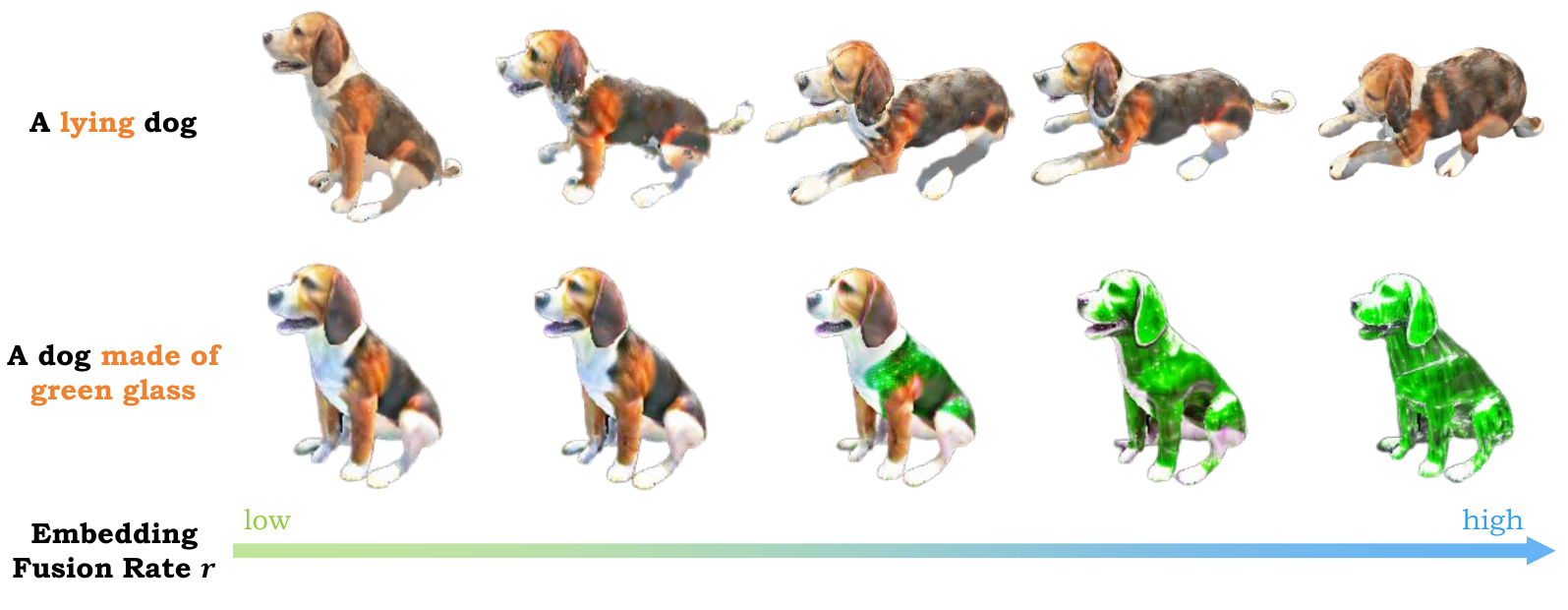}
  \caption{Controllable results of different embedding fusion rates. We show that our method can realize fine-grained control of editing extent through adjusting the embedding fusion rate $r$. As $r$ increases, the editing result\yr{s show} deeper extent to the editing target.}
  \label{fig:EF}
\end{figure}

Besides achieving high quality in non-rigid editing tasks, our method is the first to enable \yr{fine-grained} control over the granularity of edits in the semantic-driven 3D non-rigid editing task. In \yr{Fig.~}\ref{fig:EF}, we demonstrate \yr{that} adjusting the embedding fusion rate $r$ can control the intensity of edit\yr{ing, and} 
can be used to control both the geometric and appearance granularity of the edits. In the example where the target prompt is "a lying dog", as $r$ increases, the dog lies down lower. In the example where the target prompt is "a dog made of green glass", as $r$ increases, the dog's fur is gradually replaced with a green glass material.

Although theoretically the embedding fusion rate $r$ can be any number within the range of $[0,1]$, for the quality of editing, we recommend keeping $r$ within the range of $[0.35, 0.85]$.

\subsection{\yr{Details and} Hyperparameters of SPS}
\label{sec:Hyperparameters of SPS}
\yr{Our SPS introduces two additional enhancement phases: target enhancement phase, and detail enhancement phase.}
\yr{Phase} 1 \yr{(Target Enhancement)} of SPS takes the first $1,000$ steps in geometry optimization, and does not take place in texture optimization. 
In \yr{this phase} of SPS, we have Eq. (\ref{eq:7}) where $\vec{\epsilon}_{t\_e}$ is weighted by $\lambda_t$ and added to $\vec{\epsilon}_{f}$. 
In our experiment, $\lambda_t$'s value can be calculated using:
\[
\lambda_t = \frac{\Vert\vec{\epsilon}_{f}\Vert}{\Vert\vec{\epsilon}_{t\_e}\Vert} * 0.4.
\]

\yr{Phase} 3 \yr{(Detail Enhancement)} of SPS takes the last $1,000$ steps in optimization. 
In \yr{this phase} of SPS, we have Eq. (\ref{eq:8}) where $\vec{\epsilon}_{d\_e}$ is weighted by $\lambda_d$ and added to $\vec{\epsilon}_{f}$. In our experiment, $\lambda_d$'s value can be calculated using:
\[
\lambda_d = \frac{\Vert\vec{\epsilon}_{f}\Vert}{\Vert\vec{\epsilon}_{d\_e}\Vert} * 0.2.
\]


\subsection{More Experiment Results}
\label{sec:more experiment resutls}
\begin{figure}
    \centering
    \includegraphics[width=\textwidth]{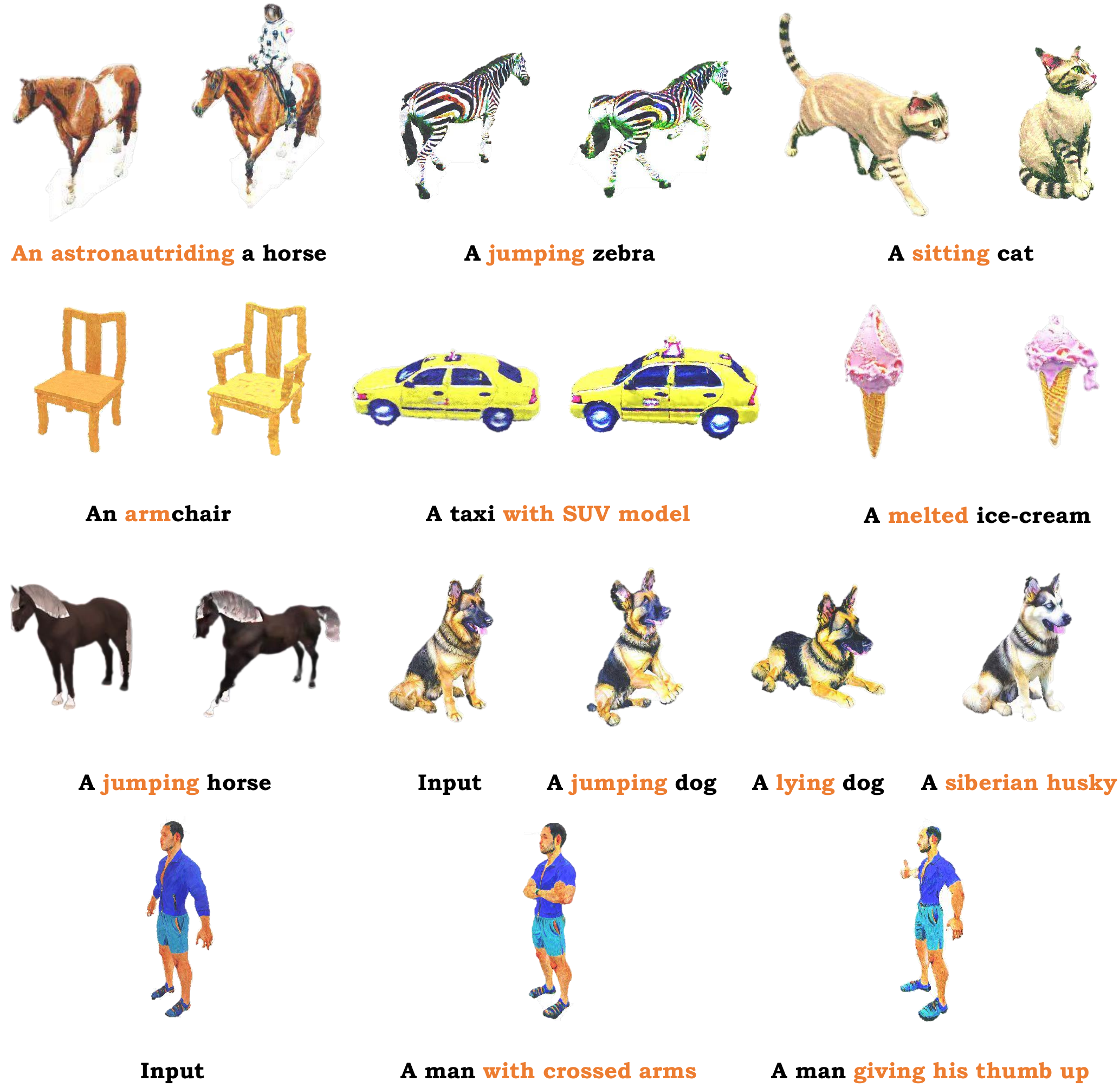}
    \caption{More experiment results of our \textit{Plasticine3D}. For each case, the original object's RGB map is displayed on the left side, with the target prompt below, highlighting the main editing objective in orange. The final edited result's RGB map is on the right side.}
    \label{fig:more_results}
\end{figure}

\hut{In this section, we provide more results of our \textit{Plasticine3D} on various 3D objects like animals, humans and cars in Fig.~\ref{fig:more_results}. It can be seen that for various \yr{original} 3D objects and target prompts, our model can \yr{consistently} achieve high-quality editing results, align\yr{ing} with the target prompt well \yr{and} keep\yr{ing} the \yr{details} of the \yr{original} 3D objects.}

\subsection{Limitations}
\label{sec:limitations}
\yr{In some scenarios, t}he scale of \yr{our edited} result\yr{s} may slightly change after geometry editing stage, so the user\yr{s} may need to resize them after the editing process. \yr{The UNet} fine-tuning \yr{step in our method} require\yr{s} a GPU device with sufficient VRAM. In order to \yr{conduct UNet fine-tuning} on consumer-grade GPU devices, gradient accumulation \yr{can be used to save memory}, 
\yr{which will lead to an increase in processing time.}
Our future work may focus on tuning-free generalized 3D editing and achieving more accurate controllable 3D editing with control points.

\subsection{Societal Impact}
\label{sec:Societal Impact}
Our \textit{Plasticine3D} has both potential positive and negative societal impacts:

\textbf{1) Positive Societal Impacts:}
\begin{itemize}
     \item \textbf{Advancement in 3D Editing Technology:} Plasticine3D introduces a novel approach to 3D non-rigid editing, which can have positive implications for various industries such as entertainment, design, and manufacturing. By enabling fine-grained control over both structure and appearance, it opens up possibilities for more creative and precise 3D modeling and editing tasks.
     \item \textbf{Enhanced Virtual Reality:} Plasticine3D offers a novel approach to 3D editing, allowing users to manipulate both the structure and appearance of virtual objects with fine-grained control. This could empower artists and designers to unleash their creativity in virtual environments, leading to the creation of innovative digital artworks, immersive experiences, and virtual worlds, which can facilitate the development of virtual reality.
\end{itemize}

\textbf{2) Negative Societal Impacts:}
\begin{itemize}
    \item \textbf{Potential for Misuse:} As with any technology that enables manipulation of digital content, there is a risk of misuse for malicious purposes such as creating deceptive or misleading visual content. Plasticine3D's capabilities, especially in terms of fine-grained control over appearance, may potentially be exploited to generate fake images or 3D models.\
    \item \textbf{Privacy Concerns:} The ability to manipulate 3D models with such precision raises concerns about privacy, particularly in scenarios where personal or sensitive information is represented in 3D form. Plasticine3D may inadvertently facilitate the creation of realistic but false representations of individuals or sensitive information, leading to concerns about privacy preservation.

\end{itemize}




\newpage
\section*{NeurIPS Paper Checklist}

\begin{enumerate}

\item {\bf Claims}
    \item[] Question: Do the main claims made in the abstract and introduction accurately reflect the paper's contributions and scope?
    \item[] Answer: \answerYes{} 
    \item[] Justification: We have clearly explained the contributions and scope in the Abstract and Introduction Section. Please refer to Page 1 and Page 2.

\item {\bf Limitations}
    \item[] Question: Does the paper discuss the limitations of the work performed by the authors?
    \item[] Answer: \answerYes{} 
    \item[] Justification: We have stated the limitations in the \yr{Appendix Sec.~\ref{sec:limitations}.} 

\item {\bf Theory Assumptions and Proofs}
    \item[] Question: For each theoretical result, does the paper provide the full set of assumptions and a complete (and correct) proof?
    \item[] Answer: \answerNA{} 
    \item[] Justification: Our paper does not include theoretical results.

    \item {\bf Experimental Result Reproducibility}
    \item[] Question: Does the paper fully disclose all the information needed to reproduce the main experimental results of the paper to the extent that it affects the main claims and/or conclusions of the paper (regardless of whether the code and data are provided or not)?
    \item[] Answer: \answerYes{} 
    \item[] Justification: The implementation details are provided in \yr{Sec.~\ref{subsec:implementation_details} and} the Appendix \yr{Sec.~\ref{sec:more implementation details}, Sec.~\ref{sec:Training Details of MVE}} and Sec.~\ref{sec:Hyperparameters of SPS}.

\item {\bf Open access to data and code}
    \item[] Question: Does the paper provide open access to the data and code, with sufficient instructions to faithfully reproduce the main experimental results, as described in supplemental material?
    \item[] Answer: \answerNo{} 
    \item[] Justification: We will open-source the code once our paper is accepted. The data comes from some open-source datasets, which are commonly used.

\item {\bf Experimental Setting/Details}
    \item[] Question: Does the paper specify all the training and test details (e.g., data splits, hyperparameters, how they were chosen, type of optimizer, etc.) necessary to understand the results?
    \item[] Answer: \answerYes{} 
    \item[] Justification: All the training and testing details can be found in \yr{Sec.~\ref{subsec:implementation_details} and} the Appendix \yr{Sec.~\ref{sec:more implementation details}, Sec.~\ref{sec:Training Details of MVE}} and Sec.~\ref{sec:Hyperparameters of SPS}.

\item {\bf Experiment Statistical Significance}
    \item[] Question: Does the paper report error bars suitably and correctly defined or other appropriate information about the statistical significance of the experiments?
    \item[] Answer: \answerNo{} 
    \item[] Justification: We have fixed the random seed\yr{s for the experiments,} and the results can be reproduced at any time. \hut{Moreover, the existing works in 3D generation usually do not report the error bars.}

\item {\bf Experiments Compute Resources}
    \item[] Question: For each experiment, does the paper provide sufficient information on the computer resources (type of compute workers, memory, time of execution) needed to reproduce the experiments?
    \item[] Answer: \answerYes{} 
    \item[] Justification: We have clearly explained the implementation details for the experiments, and reported the computation resources, which can be found in Sec.~\ref{subsec:implementation_details} and Appendix Sec.~\ref{sec:more implementation details}.
    
\item {\bf Code Of Ethics}
    \item[] Question: Does the research conducted in the paper conform, in every respect, with the NeurIPS Code of Ethics \url{https://neurips.cc/public/EthicsGuidelines}?
    \item[] Answer: \answerYes{} 
    \item[] Justification: We have carefully read NeurIPS Code of Ethics and will comply with it.

\item {\bf Broader Impacts}
    \item[] Question: Does the paper discuss both potential positive societal impacts and negative societal impacts of the work performed?
    \item[] Answer: \answerYes{} 
    \item[] Justification: We have discussed both potential positive societal impacts and negative societal impacts in Appendix Sec.~\ref{sec:Societal Impact}.
    
\item {\bf Safeguards}
    \item[] Question: Does the paper describe safeguards that have been put in place for responsible release of data or models that have a high risk for misuse (e.g., pretrained language models, image generators, or scraped datasets)?
    \item[] Answer: \answerNA{} 
    \item[] Justification: Our paper has no such risks.

\item {\bf Licenses for existing assets}
    \item[] Question: Are the creators or original owners of assets (e.g., code, data, models), used in the paper, properly credited and are the license and terms of use explicitly mentioned and properly respected?
    \item[] Answer: \answerYes{} 
    \item[] Justification: All the assets are properly cited and the licenses are properly respected.

\item {\bf New Assets}
    \item[] Question: Are new assets introduced in the paper well documented and is the documentation provided alongside the assets?
    \item[] Answer: \answerNA{} 
    \item[] Justification: Our paper does not release new assets.

\item {\bf Crowdsourcing and Research with Human Subjects}
    \item[] Question: For crowdsourcing experiments and research with human subjects, does the paper include the full text of instructions given to participants and screenshots, if applicable, as well as details about compensation (if any)? 
    \item[] Answer: \answerNA{} 
    \item[] Justification: Our paper does not involve crowdsourcing experiments nor research with human subjects.

\item {\bf Institutional Review Board (IRB) Approvals or Equivalent for Research with Human Subjects}
    \item[] Question: Does the paper describe potential risks incurred by study participants, whether such risks were disclosed to the subjects, and whether Institutional Review Board (IRB) approvals (or an equivalent approval/review based on the requirements of your country or institution) were obtained?
    \item[] Answer: \answerNA{} 
    \item[] Justification: Our paper does not involve crowdsourcing experiments nor research with human subjects.

\end{enumerate}

\end{document}